\documentclass[default,english,letterpaper, 10 pt, conference]{ieeeconf}
\usepackage[T1]{fontenc}
\usepackage[utf8]{inputenc}
\usepackage{verbatim}
\usepackage{cprotect}
\usepackage{units}
\usepackage{amsmath}
\usepackage{amssymb}
\usepackage{graphicx}

\makeatletter
\newtheorem{thm}{\protect\theoremname}

\IEEEoverridecommandlockouts

\usepackage{url}

\usepackage{cite}

\usepackage{hyperref}  
\usepackage[letterspace=100]{microtype}

\makeatother

\usepackage{babel}
\providecommand{\theoremname}{Theorem}

\begin{document}
\title{\textbf{Lyapunov-stable orientation estimator for humanoid robots}}
\author{Mehdi Benallegue, Rafael Cisneros, Abdelaziz Benallegue, \\
 Yacine Chitour, Mitsuharu Morisawa, Fumio Kanehiro \thanks{M. Benallegue, R. Cisneros, A. Benallegue, M. Morisawa, F. Kanehiro
are with the CNRS-AIST JRL (Joint Robotics Laboratory), IRL, National
Institute of Advanced Industrial Science and Technology (AIST), Tsukuba,
Japan. A. Benallegue is also Laboratoire d'Ingénierie des Systèmes
de Versailles, France, and Y. Chitour is with Université Paris-Saclay,
CentraleSupélec, CNRS, France. Email \texttt{\footnotesize{}mehdi.benallegue@aist.go.jp,
rafael.cisneros@aist.go.jp, benalleg@lisv.uvsq.fr, yacine.chitour@l2s.centralesupelec.fr,
m.morisawa@aist.go.jp, f-kanehiro@aist.go.jp }{\footnotesize{} }} }
\maketitle
\begin{abstract}
In this paper, we present an observation scheme, with proven Lyapunov
stability, for estimating a humanoid's floating base orientation.
The idea is to use velocity aided attitude estimation, which requires
to know the velocity of the system. This velocity can be obtained
by taking into account the kinematic data provided by contact information
with the environment and using the IMU and joint encoders. We demonstrate
how this operation can be used in the case of a fixed or a moving
contact, allowing it to be employed for locomotion. We show how to
use this velocity estimation within a selected two-stage state tilt
estimator: (i) the first which has a global and quick convergence
(ii) and the second which has smooth and robust dynamics. We provide
new specific proofs of almost global Lyapunov asymptotic stability
and local exponential convergence for this observer. Finally, we assess
its performance by employing a comparative simulation and by using
it within a closed-loop stabilization scheme for HRP-5P and HRP-2KAI
robots performing whole-body kinematic tasks and locomotion. 
\end{abstract}

\begin{figure}
\begin{centering}
\includegraphics[width=0.6\columnwidth]{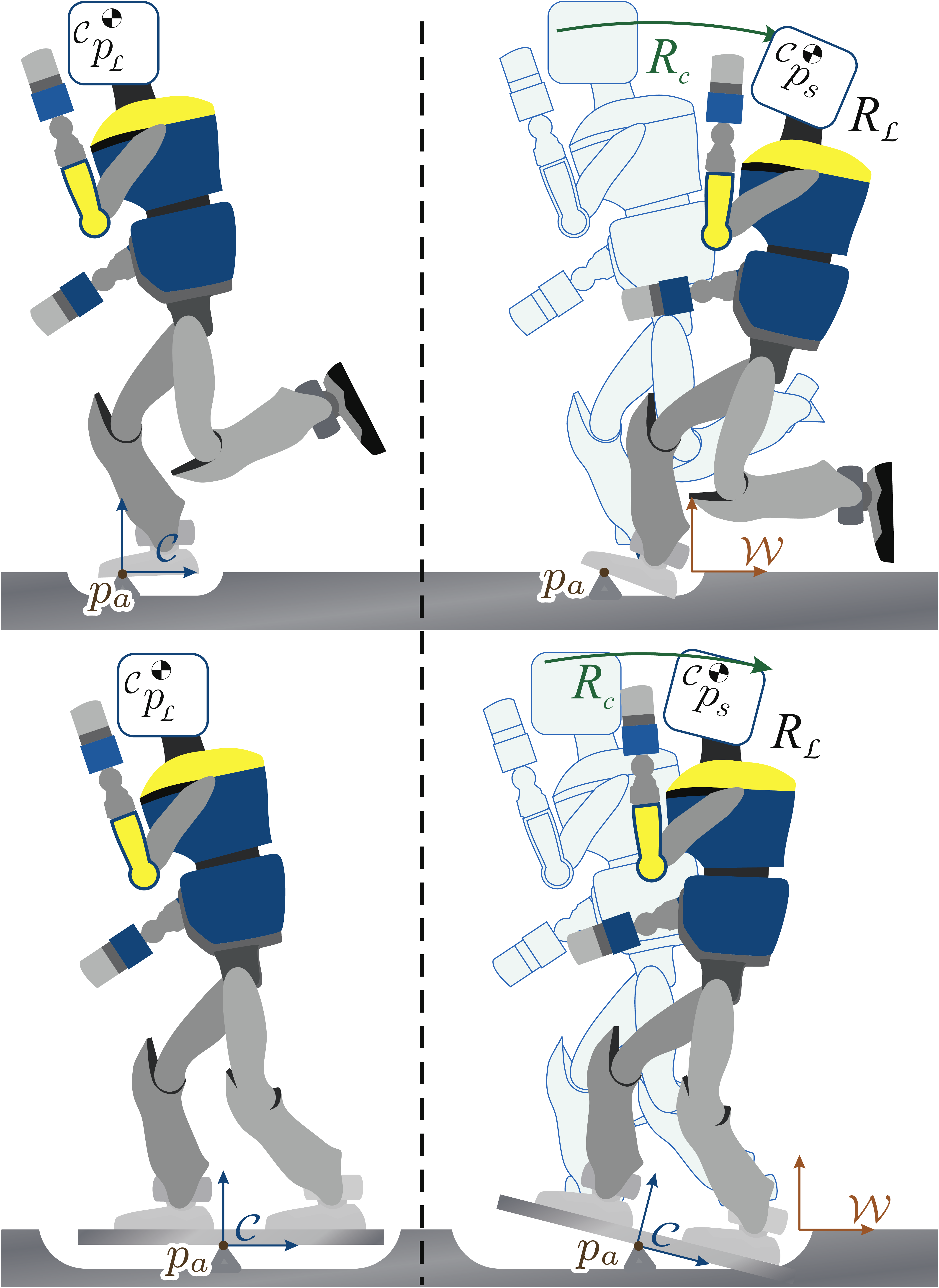} 
\par\end{centering}
\caption{Model of the anchor point definition in the case of non fixed anchor.
In the top an example of single support phase and in the bottom a
double support. The origin of the control frame $\mathcal{C}$ is
located at the position of the anchor point. The origin of the world
frame $\mathcal{W}$ can be located anywhere. The rotation $R_{\mathcal{C}}$
between the world frame $\mathcal{W}$ and the control frame $\mathcal{C}$
is not available with encoders.  }\label{fig:robot-model-moving}
\end{figure}

\section{Introduction}

A humanoid robot is a system that is intended to move through the
environment. However, this robot does not possess a dedicated actuator
for this displacement. Instead, it has to use its joint actuators
to generate contact forces with the environment, and the reaction
forces create a center-of-mass acceleration. However, these contacts
are limited by nature, mainly because of their unilaterality constraint.
This constraint implies that the feet can only push on the environment
and not pull. The consequence on a flat plane is that the center of
pressure of the contact has to lie strictly inside of the convex hull
of the support area for a motion to be physically feasible~\cite{Wieber2016}.
The estimation of the position of this center of pressure and the
prediction of its dynamics require a precise estimation of the robot
kinematics relative to the support area, especially the orientation
of a root link referred to as the floating base. In an ideal case,
this orientation could be obtained from the contact information and
the joint encoders data. However, in practice, this is not guaranteed
because of imprecise environment models and the presence of structural
compliance in the robot or the contact surfaces. Therefore, this orientation
is usually obtained using inertial measurement units (IMUs). These
sensors traditionally include gyrometers (also called gyroscopes)
and accelerometers measuring respectively, in the sensor frame, angular
velocities, and linear accelerations including the gravitational one.
Magnetometers cannot be used due to the presence of ferromagnetic
materials and powerful electric actuation. Therefore, we can only
partially reconstruct the orientation, since the yaw angle is not
observable. Only the inclination with regard to the gravitational
field, referred to as tilt, can be extracted. However, even with perfect
measurements, it is not possible to distinguish the gravitational
acceleration from the linear one using solely the sensors signals.

Several solutions to obtain this tilt have been developed. Most of
them consider that the linear accelerations are negligible and use
traditional estimators such as Kalman Filters~\cite{kajita2010iros}
or complementary filtering~\cite{allgeuer2014robust,mao2017eliminating}.
However, a humanoid robot is a system subject to large linear accelerations,
mostly due to impacts in high frequencies, structural oscillations
in middle frequencies, and bipedal locomotion in low frequencies.
Therefore, there is a need for an accurate tilt estimation that would
not be sensitive to these accelerations.

One important progress has been made in this topic by taking explicitly
into account the contact with the environment~\cite{Bloesch-RSS-12,Rotella2014,Eljaik2015Multimodal,benallegue2014humanoids,Latteur2015,wawrzynski2015ras}.
Indeed, this information provides an anchor between the robot's kinematics
and the world frame and allows to algebraically link the rotations
with the translations of the floating base. This model has also been
extended to take into account multiple pivot points due to flexibilities~\cite{vigne2018estimation},
to be merged with other sources of measurements such as LIDAR~\cite{Nobili-RSS-17,fallon2014drift}
or to take into account the dynamics in order to improve estimation
accuracy~\cite{kuindersma2016optimization}.

However, to our best knowledge, none of these methods has been shown
to give theoretically proven global (nor almost global) stability.
One important contribution was implemented, in an open-source library,
on the 2KHz loop of Cassie biped robot~\cite{hartleyIJRR2020,Hartley-RSS-18}
and was based on invariant Kalman Filtering~\cite{barrau2016invariant}
which allowed the estimation to have local stability. In our previous
paper we have shown that if we model the robot as a simple inverted
pendulum with a fixed contact point, we can use an almost global Lyapunov
stable state estimator for the tilt~\cite{Benallegue2017humanoids}.
The principle of the solution is to take advantage of the kinematic
coupling due to the contact to build algebraically a measurement of
the linear velocity expressed in the sensor frame. This additional
measurement allows us to separate the gravitational acceleration and
the linear acceleration within a simple state observer~\cite{Hua2016automatica}.
However, the limitation of this method to a fixed pivot point makes
either the solution constrained on fixed contacts or it creates discontinuities
if the contact is translated.

In this paper we extend our previous work by presenting three novel
contributions: 
\begin{enumerate}
\item The new pendulum model allows us to have a non-rigid kinematic chain
with a moving pivot point. This model then can be used for humanoid
robots during locomotion and does not create discontinuities in the
estimated velocities. 
\item We use a new state estimator with better convergence properties. This
estimator is adapted from our prior work where a more general set
of estimators has been presented~\cite{benallegue2020tac}\footnote{The paper is currently still under review, but the manuscript is available
on HAL}\textbf{.} They are based on a general two-stage observer scheme,
one having efficient exponential global convergence driving the second
one which is more robust to noise and disturbance errors. However,
in~\cite{benallegue2020tac}, only theoretical developments were
made, assuming the presence of an additional velocity measurement
and without studying their application to humanoid robots. In this
paper we extract one of these estimators, adapt it to the case of
humanoid robots, and for self-containment of the paper, we present
a new specific proof of almost global asymptotic stability and local
exponential convergence. 
\item We run simulations of a biped in a variable time-step dynamical simulator
to increase the accuracy of the simulation and test the sensitivity
of the observer to noise and modeling error of the non-rigid pendulum
model and to compare it to other methods. We also implement the estimator
on a large scale humanoid robot and perform successful closed-loop
stabilization while bending and walking. 
\end{enumerate}
The next section will formally state the problem and the following
sections will present the aforementioned contributions in the same
order. 

\section{Problem statement}

\subsection{Frames and measurements}

We denote $\mathcal{W}$ the world frame and $\mathcal{L}$ the local
frame of the sensor. To simplify notations the symbol for the world
frame $\mathcal{W}$ will be omitted. The attitude estimation has
to rely on an IMU consisting in an three-axial accelerometer and a
gyrometer. The accelerometer provides the sum of the gravitational
field and the linear acceleration of the sensor and the gyrometer
provides the angular velocity $\omega_{\mathcal{L}}$ of the IMU.
Both signals are expressed in the sensor local frame $\mathcal{L}$.
The measurements are then 
\begin{align}
y_{g}\overset{\Delta}{=} & \omega_{\mathcal{L}},\\
y_{a}\overset{\Delta}{=} & R_{\mathcal{L}}^{T}\ddot{p}_{\mathcal{L}}+g_{0}R_{\mathcal{L}}^{T}e_{z},
\end{align}
where $y_{g}$ and $y_{a}$ are the gyrometer and accelerometer measurements
given by the IMU. $R_{\mathcal{L}}$, $p_{\mathcal{L}}$ are the orientation
and the position of the IMU in the world frame. And $g_{0}$ and $e_{z}$
are respectively the standard gravity constant, and a unit vector
collinear with the gravitational field and directed upward. Finally
$\omega_{\mathcal{L}}$ is the angular velocity of the sensor expressed
in $\mathcal{L}$ such that 
\begin{equation}
\dot{R}_{\mathcal{L}}=R_{\mathcal{L}}S(\omega_{\mathcal{L}}),\label{eq:kinematics}
\end{equation}
where $S$ is the skew-symmetric operator allowing to perform cross
products.

The tilt can be entirely obtained through the estimation of the gravitational
field in $\mathcal{L}$. In other words, we need to build an estimation
of $R_{\mathcal{L}}^{T}e_{z}$.

\subsection{Linear velocity}

The estimator that we use in this paper requires an additional measurement
to the IMU, providing the vector of the linear velocity of the local
frame $\mathcal{L}$ in the world $\mathcal{W}$ expressed in $\mathcal{L}$~\cite{benallegue2020tac}.
In other words, we need an estimation of $v_{\mathcal{L}}=R_{\mathcal{L}}^{T}\dot{p}_{\mathcal{L}}$.

Note that with this measurement we can rewrite the model of the augmented
IMU as

\begin{align}
y_{v}\overset{\Delta}{=} & v_{\mathcal{L}}\\
y_{g}\overset{\Delta}{=} & \omega_{\mathcal{L}},\label{eq:mesure_IMU}\\
y_{a}\overset{\Delta}{=} & S(\omega_{\mathcal{L}})v_{\mathcal{L}}+\dot{v}_{\mathcal{L}}+g_{0}R_{\mathcal{L}}^{T}e_{z}.
\end{align}
where $y_{v}$ is the measurement of $v_{\mathcal{L}}$.

This measurement could be obtained with different kinds of sensors
such as Doppler effect sensors or by the derivation of the position
given by absolute kinematic estimators such as SLAM or LIDAR. However,
most humanoid robots are not equipped with such sensors. Nevertheless,
we show in the next section that this value can be obtained only with
contact information. 

\section{Reconstructing the linear velocity}

\subsection{The model of the contact as an anchor point}

\label{subsec:velocity-fixed-anchor}

The main specificity of legged robots is that the locomotion relies
on contacts with the environment that generate external forces. As
we said before, the contacts are not perfectly rigid and deformations
may occur. However, the contact points can be approximated to a fixed
point in the environment as long as no contact slipping occurs, and
this approximation is likely accurate to the order of the centimeter.
Since the assumption is made, the presence of an anchor point in the
environment creates a coupling between the orientation of the IMU
and its position, which gives also a relation between the angular
and the linear velocities. We show hereinafter how this coupling allows
us to provide the signal of linear velocity.

Assume we have a robot equipped with an IMU, such that there is a
perfectly-known joint-connected kinematic chain of rigid limbs linking
the IMU to the feet of the robot. There are joint encoders that can
provide the position and the velocity of each of these joints.

\begin{figure}
\begin{centering}
\includegraphics[width=0.6\columnwidth]{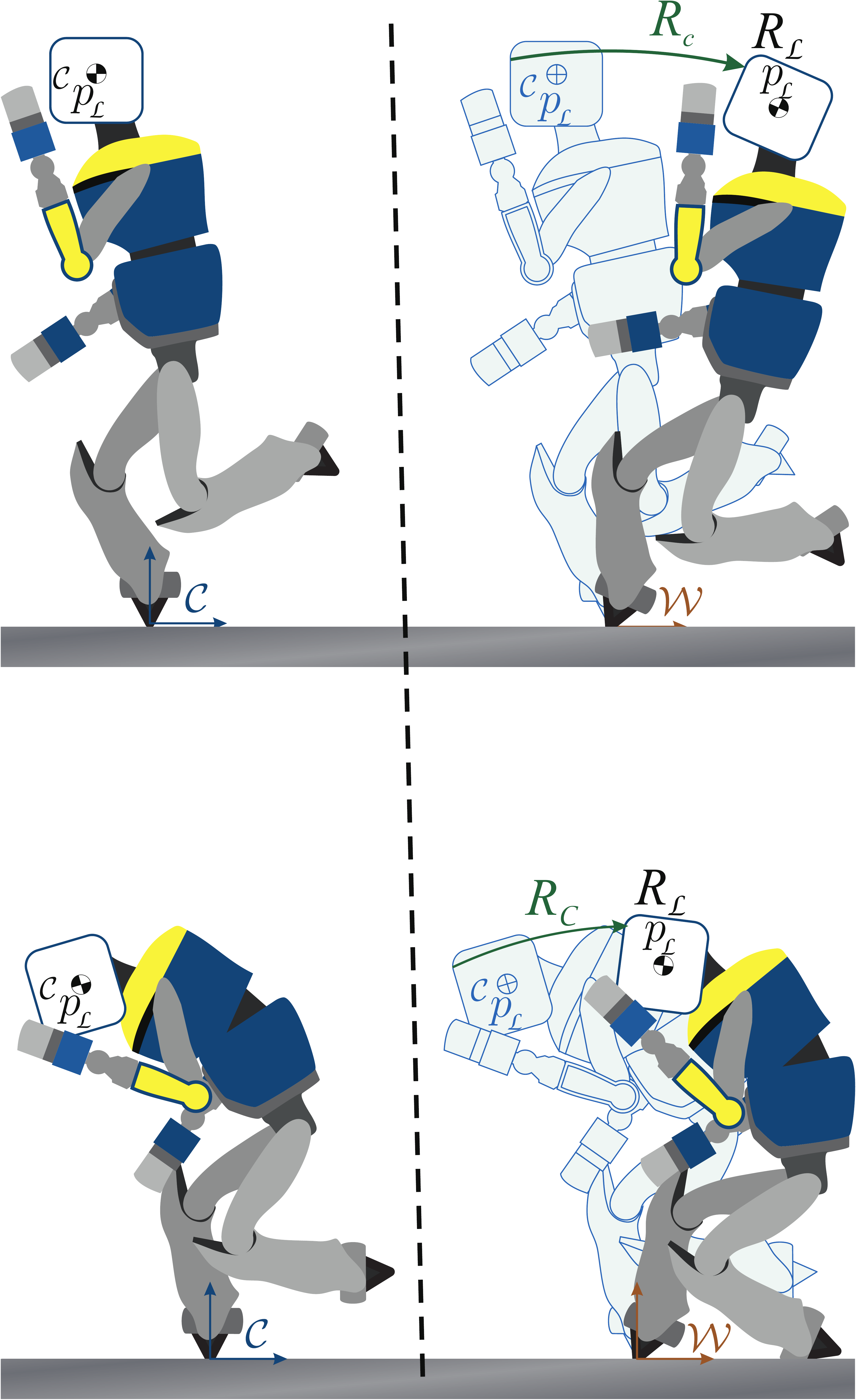} 
\par\end{centering}
\centering{}\caption{Two different configurations of the robot model, with fixed contact
point, represented on the top and on the bottom. The IMU of each robot
is represented with the circle in the head of the robot. For both
top and bottom the left figure represents the position and orientation
of the robot in the control frame $\mathcal{C}$ and the right one
represents the same robot in the world frame with the unknown rotation
$R_{\mathcal{C}}$. Note that the rotation $R_{\mathcal{C}}$ on the
top and on the bottom is the same. }
\label{fig:robot-model} 
\end{figure}

Let's take the case where contact occurs between a foot and the environment.
We assume that the contact point is at a constant position in the
environment, and we set this position at the origin of the world frame
$\mathcal{W}$ without loss of generality. If the contact and the
robot were completely rigid and the ground inclination was known,
this situation would provide us directly with the full orientation
of the IMU. This could be done thanks to the joint encoders and the
kinematic model of the robot. Indeed, simple direct kinematics would
provide the position and the orientation of the IMU as $^{\mathcal{C}}\!p_{\mathcal{L}}\in\mathbb{R}^{3}$
and $^{\mathcal{C}}\!R_{\mathcal{L}}\in SO(3)$ where $\mathcal{C}$
stands for the \emph{control} frame having its origin at the anchor
point as well and an orientation attached to any specific body of
the robot (e.g. feet). Therefore, in the ideal case of a rigid and
known environment, the control frame $\mathcal{C}$ and the world
frame $\mathcal{W}$ are identical. Furthermore, since the values
of $^{\mathcal{C}}\!p_{\mathcal{L}}$ and $^{\mathcal{C}}\!R_{\mathcal{L}}$
are obtained through the forward kinematics, their time-derivatives
are also available as $^{\mathcal{C}}\!\dot{p}_{\mathcal{L}}\in\mathbb{R}^{3}$
and $^{\mathcal{C}}\!\omega_{\mathcal{L}}\in\mathbb{R}^{3}$ such
that $^{\mathcal{C}}\!\dot{R}_{\mathcal{L}}=~^{\mathcal{C}}\!R_{\mathcal{L}}S\left(^{\mathcal{C}}\!\omega_{\mathcal{L}}\right)$
(see the left part of Figure \ref{fig:robot-model}).

However such an ideal situation does not happen and there is a difference
between the real position and orientation of the sensor in the world
frame $p_{\mathcal{L}}\in\mathbb{R}^{3}$ and $R_{\mathcal{L}}\in SO(3)$
and their values in the control frame $^{\mathcal{C}}p_{\mathcal{L}}$
and $^{\mathcal{C}}R_{\mathcal{L}}$. This difference is usually a
combination of translations and rotations, but in the absence of slipping,
we can approximate the contact point position as being fixed. This
means that the transformation between the control frame $\mathcal{C}$
and the world frame $\mathcal{W}$ is a 3D rotation $R_{\mathcal{C}}\in SO(3)$
around the origin, which means that $p_{\mathcal{L}}=R_{\mathcal{C}}~^{\mathcal{C}}\!p_{\mathcal{L}}$
and $R_{\mathcal{L}}=R_{\mathcal{C}}~^{\mathcal{C}}\!R_{\mathcal{L}}$
(see the right part of Figure \ref{fig:robot-model}).

With this model, let's develop the expression of the desired signal
of linear velocity $y_{v}$ 
\begin{align}
y_{v} & =R_{\mathcal{L}}^{T}\dot{p}_{\mathcal{L}}\\
 & =R_{\mathcal{L}}^{T}\;R_{\mathcal{C}}\;^{\mathcal{C}}\!\dot{p}_{\mathcal{L}}+R_{\mathcal{L}}^{T}\;R_{\mathcal{C}}\;S(\omega_{\mathcal{C}})~^{\mathcal{C}}\!p_{\mathcal{L}}\\
 & =~^{\mathcal{C}}\!R_{\mathcal{L}}^{T}\;^{\mathcal{C}}\!\dot{p}_{\mathcal{L}}+S(y_{g}-{}^{\mathcal{C}}\!\omega_{\mathcal{L}})~^{\mathcal{C}}\!R_{\mathcal{L}}^{T}\;^{\mathcal{C}}\!p_{\mathcal{L}}
\end{align}
which is entirely reconstructable from available signals.

\subsection{The model with moving anchor point}

\label{subsec:moving-anchor}

The model of a fixed anchor can provide the required signals to build
the attitude in the case of contact between one steady foot and the
environment. However, a humanoid robot is required to perform bipedal
walking, where the contact point transitions from one foot to the
other, with a double support phase during the transition. One solution
to this problem is to switch the contact position from one pivot point
to the other during this double support phase. However, a discontinuous
change in the pivot point generates a discontinuous $y_{v}$ signal
which introduces artificial disturbances to the estimators. Therefore,
a better solution is one where the transition is performed continuously.
In that case we assume that we know the velocity $v_{a}$ of the anchor
point $p_{a}$ in the world frame expressed in the control frame,
i.e. such that $R_{\mathcal{C}}v_{a}=\dot{p}_{a}$. This velocity
can be simply the derivative of the center of pressure reconstructed
with the force sensor or any virtual point approximating this transition.

We attach the position of the control frame origin $p_{\mathcal{C}}$
at the position of the anchor point $p_{a}$ to keep the transformation
as a pure 3D rotation around it. Note that varying this origin positions
varies the values of variables expressed in the control frame, since
they are translated by $p_{a}$. This model is sketched in Figure~\ref{fig:robot-model-moving},
where we show an example for a single and a double support. With this
new frame definition, the position of the sensor in the world frame
can be written as $p_{\mathcal{L}}=R_{\mathcal{C}}{}^{\mathcal{C}}p_{\mathcal{L}}+p_{\mathcal{C}}=R_{\mathcal{C}}{}^{\mathcal{C}}p_{\mathcal{L}}+p_{a}$.

Thanks to this expression we can develop the expression of the linear
velocity $y_{v}$ as

\begin{align}
y_{v} & =R_{\mathcal{L}}^{T}\dot{p}_{\mathcal{L}}\\
 & =R_{\mathcal{L}}^{T}\;R_{\mathcal{C}}\;^{\mathcal{C}}\!\dot{p}_{\mathcal{L}}+R_{\mathcal{L}}^{T}\;R_{\mathcal{C}}\;S(\omega_{\mathcal{C}})~^{\mathcal{C}}\!p_{\mathcal{L}}+R_{\mathcal{L}}^{T}\;R_{\mathcal{C}}\;v_{a}\\
 & =~^{\mathcal{C}}\!R_{\mathcal{L}}^{T}\;^{\mathcal{C}}\!\dot{p}_{\mathcal{L}}+S(y_{g}-{}^{\mathcal{C}}\omega_{\mathcal{L}})~^{\mathcal{C}}\!R_{\mathcal{L}}^{T}~^{\mathcal{C}}\!p_{\mathcal{L}}+{}^{\mathcal{C}}\!R_{\mathcal{L}}^{T}\;v_{a}
\end{align}
which is constructed algebraically from available signals.

Note that this formula may produce inconsistencies between the trajectory
of the anchor point $p_{a}$ and its time derivative $v_{a}=R_{\mathcal{C}}^{T}\dot{p}_{a}$.
This is because of the estimation error in the orientation of the
control frame and the lack of yaw measurements, therefore we do not
recommend using it for navigation. Nevertheless, what we need is only
a good enough approximation of the real sensor velocity $v_{\mathcal{L}}$.
We believe that the disturbance caused by this small modeling error
has much less impact than the disturbance caused by the assumption
of no linear acceleration made in classic approaches. Moreover, we
show in the next section that this measurement is used through low-pass
processing within a nonlinear complementary filter, to produce a reliable
and theoretically grounded tilt estimation. 

\section{Tilt estimation }\label{sec:Tilt-estimation}

\subsection{State and measurement definitions}

Let's define the following state variables 
\begin{eqnarray}
x_{1} & \overset{\Delta}{=} & v_{\mathcal{L}},\label{eq:x1}\\
x_{2} & \overset{\Delta}{=} & R_{\mathcal{L}}^{T}e_{z},\label{eq:x2}
\end{eqnarray}
with $x_{1}\in\mathbb{R}^{3}$ and $x_{2}\in\mathbb{S}^{2}$ , where
the set $\mathbb{S}^{2}\subset\mathbb{R}^{3}$ is the unit sphere
centered at the origin, and defined as 
\[
\mathbb{S}^{2}=\left\{ x\in\mathbb{R}^{3}|\left\Vert x\right\Vert =1\right\} .
\]

The variable $x_{1}$ is considered known using $y_{v}$, even if
noisy. On the contrary, $x_{2}$ is the tilt we aim to estimate and
cannot be obtained algebraically from the measurements.

From equations (\ref{eq:mesure_IMU}) and (\ref{eq:x1}) we get 
\begin{align}
\dot{x}_{1}= & -S(y_{g})x_{1}-g_{0}R_{\mathcal{L}}^{T}e_{z}+y_{a}.\label{eq:x1-dynamics}
\end{align}

This, together with the time-differentiation of $x_{2}$, provide
us with the following state dynamic equations

\begin{equation}
\begin{cases}
\dot{x}_{1} & =-S(y_{g})x_{1}-g_{0}x_{2}+y_{a},\\
\dot{x}_{2} & =-S(y_{g})x_{2}.
\end{cases}\label{eq:dynamics}
\end{equation}

The system (\ref{eq:dynamics}) is suitable for the observer synthesis
presented hereinafter.

\subsection{Two-stage observer designed in $\mathbb{R}^{3}\times\mathbb{R}^{3}\times\mathbb{S}^{2}$
and error dynamics }

\label{subsec:State-observer-in-R3xR3xS2}

We define the observer for the state defined in Equations (\ref{eq:x1})
and (\ref{eq:x2}). It is designed in $\mathbb{R}^{3}\times\mathbb{R}^{3}\times\mathbb{S}^{2}$
and is given by 
\begin{equation}
\begin{cases}
\dot{\hat{x}}_{1} & =-S\left(y_{g}\right)\hat{x}_{1}-g_{0}\hat{x}'_{2}+y_{a}+\alpha_{1}\left(y_{v}-\hat{x}_{1}\right)\\
\dot{\hat{x}}'_{2} & =-S\left(y_{g}\right)\hat{x}'_{2}-\frac{\alpha_{2}}{g_{0}}\left(y_{v}-\hat{x}_{1}\right)\\
\dot{\hat{x}}_{2} & =-S\left(y_{g}-\gamma S\left(\hat{x}_{2}\right)\hat{x}'_{2}\right)\hat{x}_{2}
\end{cases}\label{eq:RxRxS_observer}
\end{equation}
where $\alpha_{1}$, $\alpha_{2}$ and $\gamma$ are positive scalar
gains, $\hat{x}_{1}$ and $\hat{x}_{2}$ are estimations of $x_{1}$
and $x_{2}$ respectively and $\hat{x}'_{2}$ is an intermediate estimation
of $x_{2}$.

Assuming perfect measurements and using the estimation errors defined
as $\tilde{x}_{1}\overset{\Delta}{=}x_{1}-\hat{x}_{1}$, $\tilde{x}'_{2}\overset{\Delta}{=}x_{2}-\hat{x}'_{2}$
and $\tilde{x}_{2}=x_{2}-\hat{x}_{2}$, we get the error dynamics
as

\begin{equation}
\begin{cases}
\dot{\tilde{x}}_{1} & =-S(y_{g})\tilde{x}_{1}-\alpha_{1}\tilde{x}_{1}-g_{0}\tilde{x}'{}_{2}\\
\dot{\tilde{x}}'_{2} & =-S(y_{g})\tilde{x}'_{2}+\frac{\alpha_{2}}{g_{0}}\tilde{x}_{1}\\
\dot{\tilde{x}}_{2} & =-S(y_{g})\tilde{x}_{2}+\gamma S^{2}(\hat{x}_{2})\tilde{x}_{2}-\gamma S^{2}(\hat{x}_{2})\tilde{x}'_{2}
\end{cases}\label{eq:RxRxS_error_dynamics}
\end{equation}

To run the analysis of errors, we set $z_{i}=R_{\mathcal{L}}\tilde{x}_{i}$
and $z'_{2}=R_{\mathcal{L}}\tilde{x}'{}_{2}$. Noticing $R_{\mathcal{L}}\hat{x}_{2}=e_{z}-z_{2}$,
one gets 
\begin{equation}
\begin{cases}
\dot{z}_{1} & =-\alpha_{1}z_{1}-g_{0}z'_{2}\\
\dot{z}'_{2} & =\frac{\alpha_{2}}{g_{0}}z_{1}\\
\dot{z}_{2} & =\gamma S^{2}(e_{z}-z_{2})\left(z_{2}-z'_{2}\right)
\end{cases}\label{eq:RxRxS_error_dynamics_2}
\end{equation}
This new dynamics is autonomous and defines a time-invariant ordinary
differential equation (ODE) which simplifies the stability analysis.
Indeed, if one defines the error state $\xi\overset{\Delta}{=}\left(z_{1},z'_{2},z_{2}\right)$
and the state space $\varUpsilon:=\mathbb{R}^{3}\times\mathbb{R}^{3}\times\mathbb{S}_{e_{z}}$
with $\mathbb{S}_{e_{z}}=\left\{ z\in\mathbb{R}^{3}|\left(e_{z}-z\right)\in\mathbb{S}^{2}\right\} $,
one can write (\ref{eq:RxRxS_error_dynamics_2}) as $\dot{\xi}=F\left(\xi\right)$
where $F$ gathers the right-hand side of (\ref{eq:RxRxS_error_dynamics_2})
and defines a smooth vector field on $\varUpsilon$.

Note that the first two lines of (\ref{eq:RxRxS_observer}) constitute
a separate tilt estimator defined in $\mathbb{R}^{3}\times\mathbb{R}^{3}$.
A comparable tilt estimator can be found in \cite{Martin2016arxiv}.
We show hereinafter the convergence and the performances of this intermediate
estimation. We show, afterward, that even if this intermediate estimator
has very good performances, the estimator of (\ref{eq:RxRxS_observer})
is an extension allowing to ensure that the tilt estimation respects
the normality constraint and reject more disturbances.

\subsubsection{Global exponential stability of the intermediate estimator}

\label{subsec:RxR}

The two first lines of the observer given by equation (\ref{eq:RxRxS_observer})
constitute an independent observer of the tilt which is designed in
$\mathbb{R}^{3}\times\mathbb{R}^{3}$ where the vector $\hat{x}'_{2}\in\mathbb{R}^{3}$
is the estimation of $x_{2}=R_{\mathcal{L}}^{T}e_{z}$. In this case
the error dynamics is given by 
\begin{equation}
\begin{cases}
\dot{z}_{1} & =-\alpha_{1}z_{1}-g_{0}z'_{2}\\
\dot{z}'_{2} & =\frac{\alpha_{2}}{g_{0}}z_{1}
\end{cases}\label{eq:error_dynamics_in_R3}
\end{equation}
The dynamics is linear and autonomous. If one define the state $\xi^{\prime}:=\left(\nicefrac{\alpha_{2}}{g_{0}}z_{1},z'_{2}\right)$
and the state space $\varUpsilon':=\mathbb{R}^{3}\times\mathbb{R}^{3}$,
one can write (\ref{eq:error_dynamics_in_R3}) as $\dot{\xi}^{\prime}=A_{1}\xi^{\prime}$
where $A_{1}$ is constant and given by 
\begin{equation}
A_{1}=\left[\begin{array}{ccc}
-\alpha_{1}I_{3\times3} &  & -\alpha_{2}I_{3\times3}\\
I_{3\times3} &  & 0
\end{array}\right]
\end{equation}

This matrix allows us to state the following theorem.
\begin{thm}
\textup{\emph{The system (\ref{eq:error_dynamics_in_R3}) is globally
exponentially stable with respect to the origin $(0,0)$.}}

The proof comes from the stability of the matrix $A_{1}$.
\end{thm}
The intermediate estimation $\hat{x}'_{2}$ is not constrained to
be normalized, despite $x_{2}=R_{\mathcal{L}}^{T}e_{z}$ being a unit
vector. This constitutes a strength of this estimator, it will converge
through the unit sphere from any initial value. This constitutes also
two weaknesses. The first one is that a simple normalization of this
estimation would risk undefined output and unbounded time-derivatives
when the norm is close to zero. The second one is the sensitivity
noise including the one which leads to the violation of normality
constraint. The solution we propose is to add the third line of (\ref{eq:RxRxS_observer})
to take profit from the convergence of $\hat{x}'_{2}$ by tracking
it while maintaining the normality constraint of the tilt estimation.

\subsubsection{Almost global asymptotic stability of the full estimator}

Let's now consider the full estimator of Equation~(\ref{eq:RxRxS_observer}).
We state the following theorem,
\begin{thm}
\textup{\emph{\label{thm:RxRxS_asympt_conv}The time-invariant ODE
defined by (\ref{eq:RxRxS_error_dynamics_2}) is almost globally stable
with respect to the origin in the following sense: there exists an
open dense set $\varUpsilon_{0}\subset\varUpsilon$ such that, for
every initial condition $\xi_{0}\in\varUpsilon_{0}$, the corresponding
trajectory converges asymptotically to ($0,0,0$).}} 
\end{thm}
The proof is presented in the appendix.

In addition to this asymptotic stability, we can also show good local
performances around the desired equilibrium. It is indeed really important
for humanoid robots to maintain precise tilt observation after the
convergence of the estimation since it is involved in high gain closed-loop
stabilization.

\subsubsection{Local exponential stability}

The good local performances of this estimator can be stated with the
following theorem 
\begin{thm}
\textup{\emph{\label{thm:RxRxS_exponential}The time-invariant ODE
defined by (\ref{eq:RxRxS_error_dynamics_2}) provides local exponential
convergence of the error $\xi$ to $\left(0,0,0\right)$.}} 
\end{thm}
The proof is presented in the appendix.

These theoretical guarantees are provided in the presence of precise
measurements. But since the sensors are noisy and the velocity measurement
has been produced with an approximation, we need to assess the performances
and compare them with state-of-the-art approaches. Therefore in the
next section, we present the implementation of this observer on a
dynamical simulation and on the full-size humanoid robots HRP-5P and
HRP-2KAI. 

\section{Implementation}

\subsection{Simulations}

\begin{figure}
\begin{centering}
\includegraphics[width=1\columnwidth]{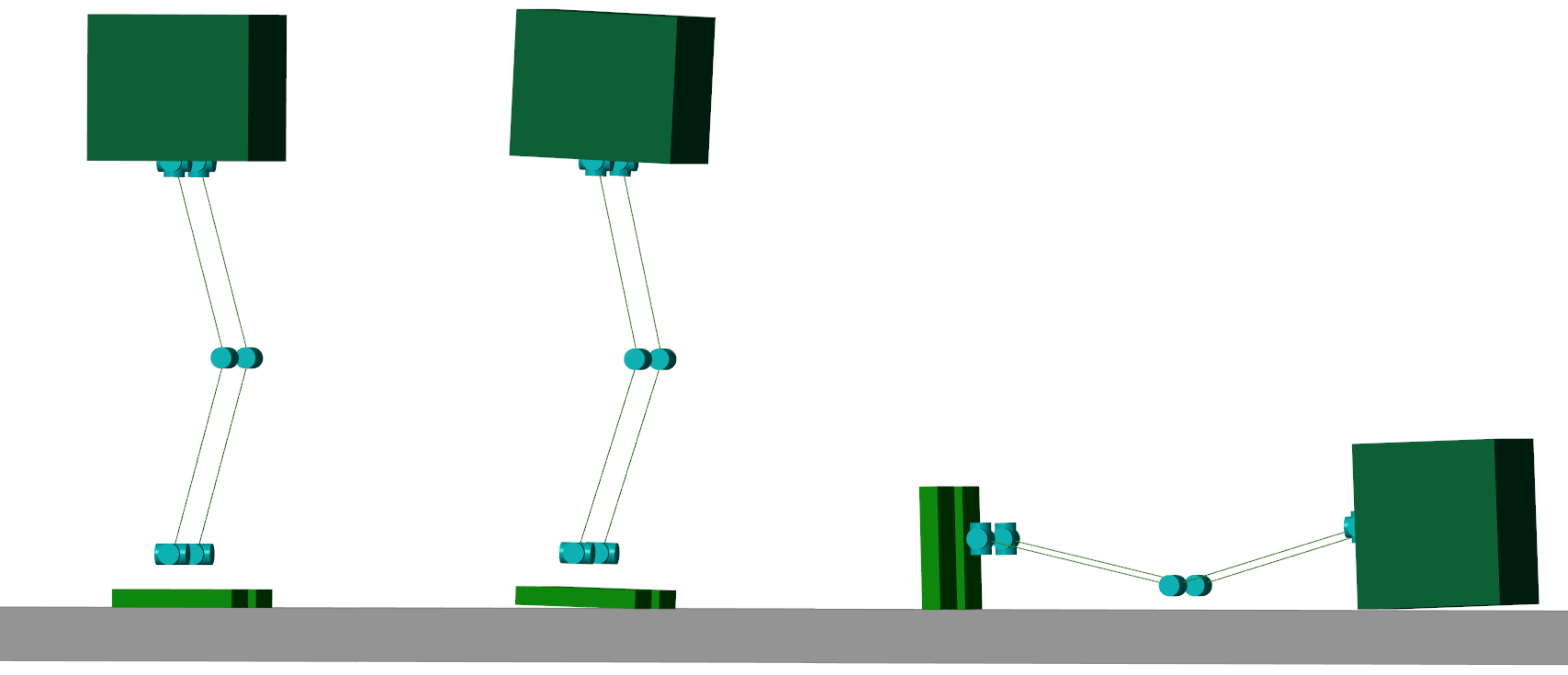} 
\par\end{centering}
\caption{Screenshots of the simulation of the simple biped. The first image
showing it in rest position, the second one after the small push and
the third one after the big push. }
\label{fig:simple-biped} 
\end{figure}

The approach validity can be assessed through a precise simulation.
It is performed under the Matlab Simscape Multibody environment and
uses a model of a simple 12 degrees of freedom biped robot weighing
42.6Kg with two rectangular feet. Figure \ref{fig:simple-biped} shows
simulation screen captures. The contact with the environment is simulated
with point contact at the four corners of each foot. With the effect
of the weight, each contact point gets strictly inside the ground
and generates a viscoelastic reaction force to simulate a compliant
environment with stiffness of 10$^{5}$ N/m and damping of 10$^{3}$
$\nicefrac{\text{N}s}{\text{m}}$. These contact dynamics change the
orientation of the contact when the forces change. The robot is set
to rest configuration, with a little oscillation during convergence,
and it is pushed twice, at 4 s and 14 s with 100 N and 300 N forces
respectively, each lasting 0.1 s. The first one leads the robot to
tilt and oscillate until convergence and the second makes the robot
tip and fall. Note that this oscillation does not have a fixed point
in the environment and therefore breaks assumptions of a fixed anchor
point.

The signals of an IMU consisting only of a gyrometer and an accelerometer
were simulated with a uniform Gaussian noise with SD of 0.02 rad/s
and 0.5 m/s$^{2}$ respectively. We use the position of the ZMP as
a position of the anchor point. This allows us to generate the signal
of the velocity $x_{1}$ using the method of Section \ref{subsec:moving-anchor}.
Three state estimators were tested: (i) our estimator in $\mathbb{R}^{3}\times\mathbb{R}^{3}\times\mathbb{S}^{2}$
of Sec. \ref{subsec:State-observer-in-R3xR3xS2}, (ii) the estimator
in $\mathbb{R}^{3}\times\mathbb{S}^{2}$ presented in our previous
work~\cite{Benallegue2017humanoids}, and the estimator of Hua 2016
from \cite{Hua2016automatica}, which is also based on velocity-aided
IMU and is comparable with our previous work, but with different dynamic
properties. The parameters of the three estimators were $\alpha_{2}=k_{1}^{r}=20$,
$\alpha_{1}=k_{1}^{v}=k_{2}^{v}=100$ and $\gamma=3$ and they were
initialized to an error of 0.2 rad around the $X$ axis. The simulations
are very slow (about 25 minutes) because of the variable time-step
duration allowing for high precision in the simulation.

\begin{figure}
\begin{centering}
\includegraphics[width=0.75\columnwidth]{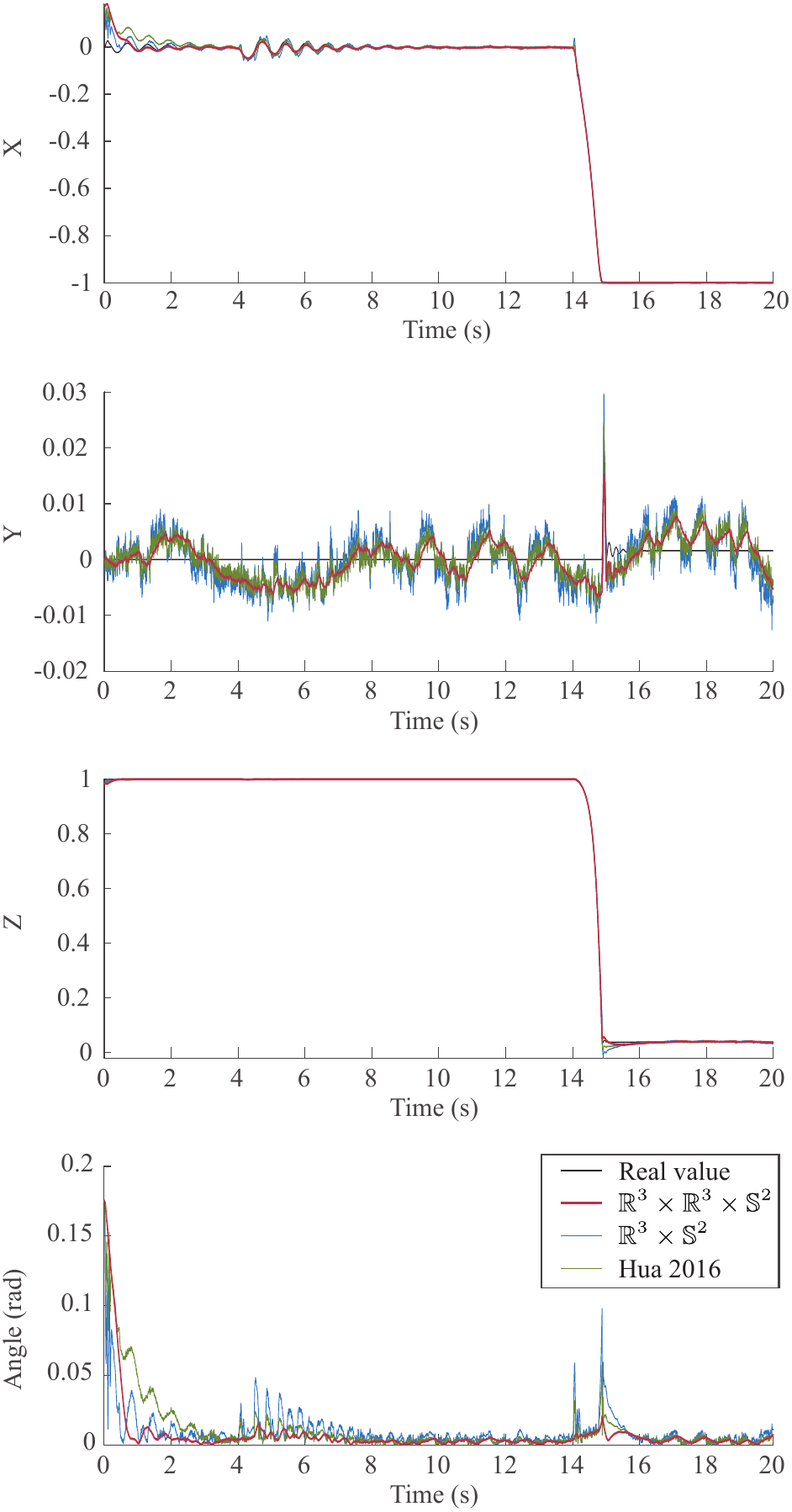}
\par\end{centering}
\caption{Plot showing the evolution of estimations from three tilt observers
for the simulated biped. The top three figures show each component
of the tilt estimation $\hat{x}_{2}$ compared to the real value.
The bottom shows the error angle for the three estimators. }
\label{fig:plot-simple-biped} 
\end{figure}

The plots of Figure \ref{fig:plot-simple-biped} show the result of
this simulation. The beginning includes the convergence phase of the
estimators. We note that our estimator in $\mathbb{R}^{3}\times\mathbb{R}^{3}\times\mathbb{S}^{2}$
is the fastest to converge and the estimator of Hua 2016 is the slowest.
The external forces are then applied at 4s where we can see (in the
bottom) that the error has increased. This is only due to the modeling
error arising from the assumption of an anchor point. However, by
comparing the actual tilt with the estimation we can see that the
tracking quality is good, especially the estimator in $\mathbb{R}^{3}\times\mathbb{R}^{3}\times\mathbb{S}^{2}$.
Interestingly the estimator of Hua 2016 has good robustness to this
modeling error compared to the estimator in $\mathbb{R}^{3}\times\mathbb{S}^{2}$.
This could be specifically due to the slowness of Hua's estimator
providing it with stronger filtering of disturbances than our previous
method. But the estimation of our method in $\mathbb{R}^{3}\times\mathbb{R}^{3}\times\mathbb{S}^{2}$
is still better. However, the most interesting part is the fall induced
at 14s where all estimators have rather good tracking thanks to their
convergence properties. Nevertheless, If we focus on the error on
the bottom, we can see that the previous estimators have an increased
error while our estimator did not suffer a lot from this important
disturbance. It is interesting to observe also, in the estimation
on the Y axis, the noise sensitivity of the methods. We can see that
our method in $\mathbb{R}^{3}\times\mathbb{R}^{3}\times\mathbb{S}^{2}$
has the best filtering, followed by Hua 2016 and then our previous
method in $\mathbb{R}^{3}\times\mathbb{S}^{2}$.

These results confirm the good behavior of the estimator in $\mathbb{R}^{3}\times\mathbb{R}^{3}\times\mathbb{S}^{2}$
which is the fastest to converge without being too sensitive to noises
and disturbances.

\subsection{Open-source implementation on HRP-5P and HRP-2KAI}

The estimator in $\mathbb{R}^{3}\times\mathbb{R}^{3}\times\mathbb{S}^{2}$,
coupled with the linear velocity reconstruction exposed in the previous
section has been implemented in the control framework of two robots,
HRP-5P~\cite{kaneko2019humanoid} and HRP-2KAI~\cite{kaneko2015humanoid}.
Both are full-scale humanoid robots of 182 cm, 101 kg and 170cm, 65Kg
respectively, and are equipped with 37 and 32 degrees of freedom,
developed at the National Institute of Advanced Industrial Science
and Technology (AIST). Both robots are controlled in a real-time framework
based on Open-RTM middleware developed in AIST as well. Nevertheless,
the implementation of the tilt estimator has rather been performed
as a standalone state observation open-source C++ library \cite{tiltEstimatorGithub},
called simply tilt estimator.

The tilt estimation is used in the real-time balance control of the
robots described in \cite{Morisawa_SII2014}, which requires an estimation
of the position and the velocity of the center of mass. To do that,
let us assume that the positions of the feet are known. They can be
for example provided by a footstep planner. Note that although a precise
location is required for the navigation, it is not necessary for the
balance control. From the feet position, the control frame $\mathcal{C}$
is built. The anchor $p_{a}$, at the origin of $\mathcal{C}$, can
be described by a weighted ratio of the feet locations, the weights
being the desired vertical forces at each foot. This definition allows
us to keep a continuous trajectory of the origin while successfully
transiting from one foot to another, and to know the velocity of this
origin $v_{a}$. The position $^{\mathcal{C}}p_{\mathcal{L}}$ and
orientation $^{\mathcal{C}}R_{\mathcal{L}}$ of the IMU in $\mathcal{C}$
are known from the model and the encoders, and the linear and angular
velocities $^{\mathcal{C}}\dot{p}_{\mathcal{L}}$ and $^{\mathcal{C}}\omega_{\mathcal{L}}$
can be estimated by using finite differences. This allows implementing
the estimator as described in Section \ref{subsec:moving-anchor}.
The desired yaw coming from the bipedal locomotion planner is combined
with the estimated tilt using the TRIAD method~\cite{shuster1981jgc},
where we construct a virtual magnetometer by using the desired yaw
angle without disturbing the estimation of the tilt. This attitude
estimation provides us with the orientation of the floating-base,
for instance, the waist, and the position of the anchor point gives
us the floating-base position. The angular and linear velocities of
the base are obtained with the gyrometer and the anchor point. Having
estimated the state of the floating-base it is straightforward to
estimate the state of the center of mass. This is the necessary feedback
that is required to implement the balance control.

\begin{figure}
\begin{centering}
\includegraphics[width=0.9\linewidth]{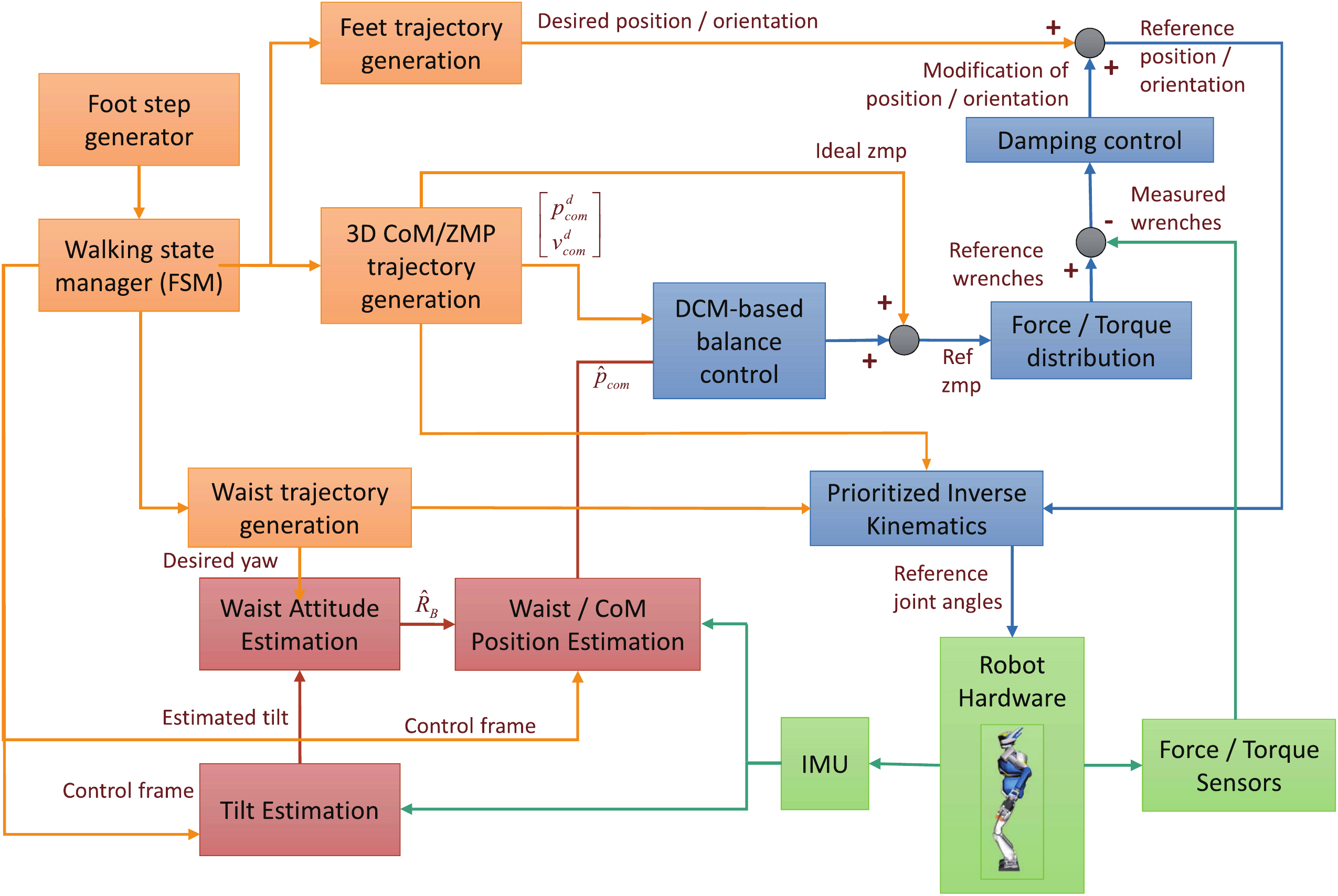}
\par\end{centering}
\caption{A simplified diagram of the bipedal locomotion control framework.
Elements in red are components related to the estimation. Elements
in yellow are components used to generate desired values. Elements
in blue are components related to the control. Elements in green belong
to the robot. }
\label{fig:Diagram} 
\end{figure}

The balance control is realized by a PID controller of the error between
the desired and the estimated Divergent Component of Motion (DCM).
The DCM is calculated as $\xi=c+\frac{1}{\rho_{c}}\dot{c}$, where
$c$ and $\dot{c}$ are the CoM position and its velocity, and $\rho_{c}$
is the natural frequency of the dynamics of the equivalent inverted
pendulum~\cite{Englsberger_IROS2013}. The DCM-based controller modifies
the reference center of pressure, from which the reference forces
and moments of the feet are computed. Then, a damping control, described
in \cite{kajita2010iros}, modifies the relative position of the feet
and their orientation to realize the foot reference force and moment.
Finally, the joint angles are calculated by using the prioritized
inverse kinematics scheme proposed by \cite{Kanoun_TRO2011} and sent
to the joint servos. \figurename~\ref{fig:Diagram} illustrates
the previous description.

\begin{figure}
\includegraphics[width=1\columnwidth]{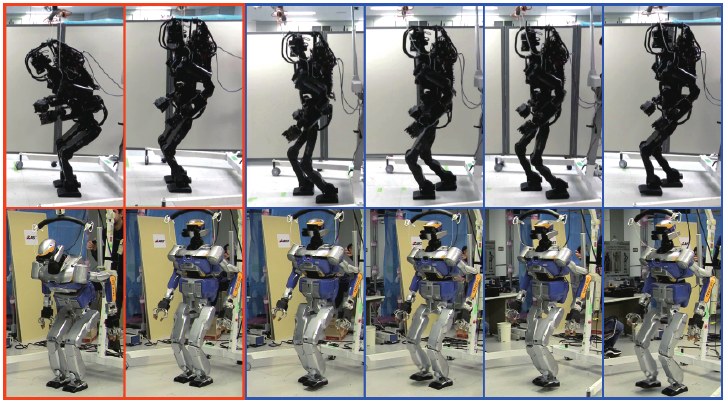}

\caption{Screen capture of the robots (HRP-5P on top and HRP-2KAI on the bottom)
performing the test motions. The two leftmost images show the leaning
task and the remaining images show the walking motion. }
\label{fig:capture} 
\end{figure}

\begin{figure}
\begin{centering}
\includegraphics[width=1\columnwidth]{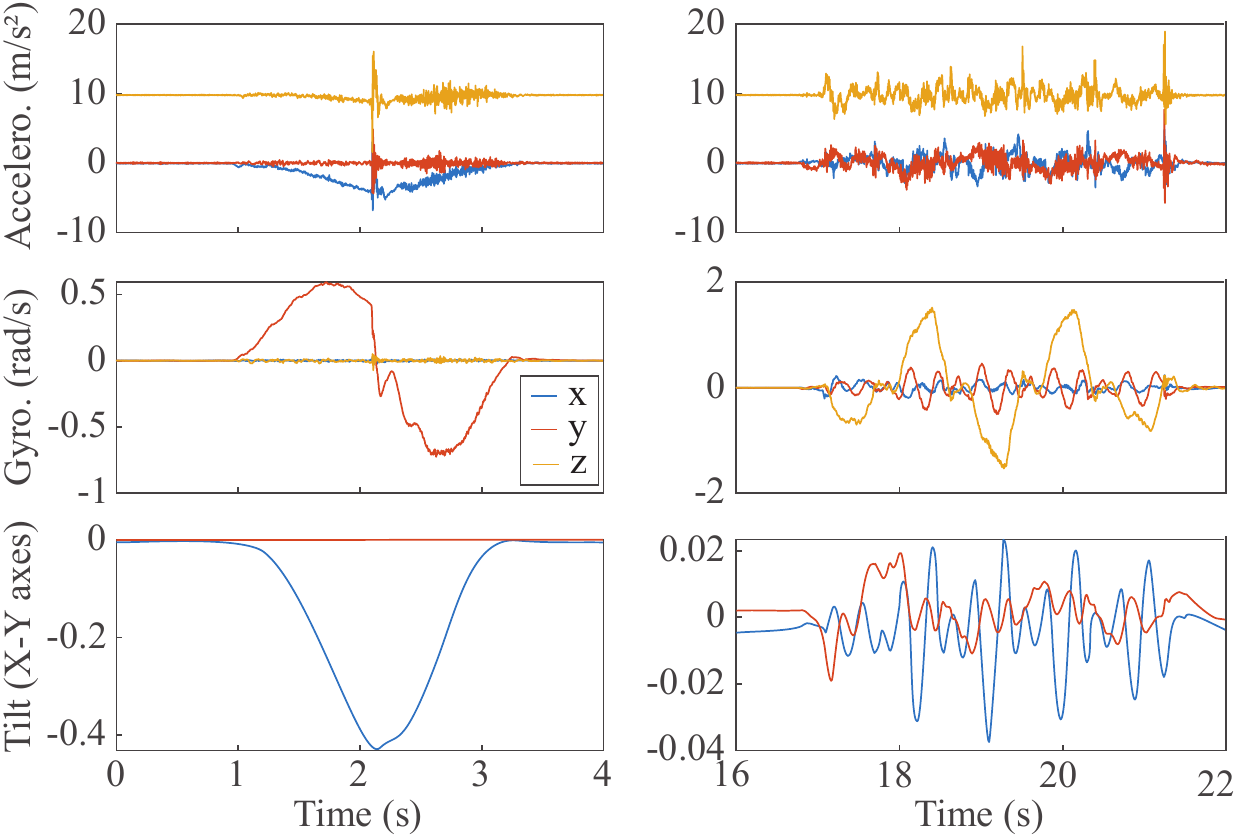}
\par\end{centering}
\caption{Plots showing the chest leaning test on the left and the walking test
on the right. The accelerometer measurements are on the top, the gyrometer
measurements on the middle, and the tilt estimation on the bottom.
The Z axis has been removed from the tilt estimation for legibility
and can be reconstructed knowing that it is the positive solution
to the vector normality constraint. }
\label{fig:realplots} 
\end{figure}

Since there is no ground truth value for the real robot orientation,
we test the performance of the estimator by assessing the stability
in two experiments. The first one consists of leaning the chest forward
by 25 degrees and coming back to the vertical, all in 2 seconds. This
allows us to check the reactivity of the estimator to fast tilt changes,
similarly to what happens in the falling simulation above. The second
test is walk forward, this will assess the behavior concerning impacts
and modifications of the contact point. All these experiments were
successful and the balance of both robots could be maintained. Screen
captures of these tests are available in Figure \ref{fig:capture}.
For the case of HRP-5P, we show, in Figure (\ref{fig:realplots}),
the measurements of the IMU together with the tilt estimation during
each experiment. There we can see the tracking of the inclination
of the chest and the quality of estimation during walking. We can
evaluate the gait impacts by looking at the accelerometer measurements.

\section{Conclusion}

\label{sec:Discussion-and-conclusion}

We have presented a Lyapunov stable tilt estimator that can be applied
to humanoid robots. We have shown how the presence of an anchor position,
even if it is moving, can be exploited to algebraically produce approximations
of the linear velocity of the sensor. This new signal allows us to
use estimators dedicated to velocity-aided IMUs. We have shown the
theoretical guarantees of almost global asymptotic stability and local
exponential convergence of the presented two-stage estimator. Finally,
we have studied the performance of the estimator on a dynamic simulator
of a simple biped and have shown the stability of real robots when
used within a closed-loop balance control framework.

\appendix

\section{Proofs}

\subsection{Proof of theorem \ref{thm:RxRxS_asympt_conv}}
\begin{proof}
We define the positive-definite differentiable function $V:\varUpsilon\rightarrow\mathbb{R}^{+}$
radially unbounded over $\varUpsilon$ 
\begin{align}
V & \overset{\Delta}{=}\frac{\alpha_{2}}{8\alpha_{1}g_{0}}\left\Vert z_{1}\right\Vert {}^{2}+\frac{1}{8\alpha_{1}g_{0}}\left\Vert \alpha_{1}z_{1}+g_{0}z'_{2}\right\Vert {}^{2}\nonumber \\
 & \qquad+\frac{\alpha_{1}g_{0}}{4\alpha_{2}}\left\Vert z'_{2}\right\Vert {}^{2}+\frac{1}{2\gamma}\left\Vert z_{2}\right\Vert {}^{2}.\label{eq:lyap_2}
\end{align}

Using the error dynamics given by (\ref{eq:RxRxS_error_dynamics}),
the time derivative of $V$ is then given by 
\begin{align}
\dot{V} & =-\frac{\alpha_{1}^{2}}{4g_{0}}z_{1}^{T}z_{1}-\frac{g_{0}}{4}z_{2}^{\prime T}z{}_{2}^{\prime}+\frac{1}{\gamma}z_{2}^{T}\dot{z}_{2},\\
\dot{V} & =-\frac{\alpha_{1}^{2}}{4g_{0}}z_{1}^{T}z_{1}-\frac{g_{0}}{4}z{}_{2}^{\prime T}z{}_{2}^{\prime}+z_{2}^{T}S^{2}(e_{z})z_{2}\nonumber \\
 & \qquad-z_{2}^{T}S(e_{z})S(e_{z}-z_{2})z'_{2}\label{eq:lyap_v2_dot}
\end{align}

If we define the new vector $w=S(e_{z})z_{2}$, we can write 
\[
\dot{V}=-\frac{\alpha_{1}^{2}}{4g_{0}}z_{1}^{T}z_{1}-\frac{g_{0}}{4}z_{2}^{\prime T}z{}_{2}^{\prime}-w^{T}w+w^{T}S(e_{z}-z_{2})z'_{2}
\]
and then 
\[
\dot{V}\leq-\frac{\alpha_{1}^{2}}{4g_{0}}\left\Vert z_{1}\right\Vert ^{2}-\frac{g_{0}}{4}\left\Vert z'_{2}\right\Vert ^{2}-\left\Vert w\right\Vert ^{2}+\left\Vert w\right\Vert \left\Vert z'_{2}\right\Vert 
\]

We define the vector $\varrho=[\begin{array}{ccc}
\left\Vert z_{1}\right\Vert  & \left\Vert z'_{2}\right\Vert  & \left\Vert w\right\Vert \end{array}]^{T}$, giving 
\[
\dot{V}\leq-\varrho^{T}H\varrho
\]
where $H=\left[\begin{array}{ccccc}
\frac{\alpha_{1}^{2}}{4g_{0}} & 0 & 0\\
0 & \frac{g_{0}}{4} & -\frac{1}{2}\\
0 & -\frac{1}{2} & 1
\end{array}\right]$, a positive definite matrix. It means that $\dot{V}\leq0$ and more
specifically $\dot{V}<0$ if $(z_{1},z'{}_{2},z_{2})$ is not an equilibrium
point.

The linearized system around the equilibrium $(0,0,2e_{z})$ is given
by the following dynamics 
\begin{equation}
\dot{X}=A_{2}X,
\end{equation}
with $X=\left[\begin{array}{ccc}
\left(z_{1}\right)^{T} & \left(z'_{2}\right)^{T} & \left(z_{2}-2e_{z}\right)^{T}\end{array}\right]^{T}$ and $A_{2}$ being a constant matrix having the form 
\begin{equation}
A_{2}=\left[\begin{array}{ccc}
-\alpha_{1}I_{3\times3} & -g_{0}I_{3\times3} & 0\\
\frac{\alpha_{2}}{g_{0}}I_{3\times3} & 0 & 0\\
0 & -\gamma S^{2}(e_{z}) & -\gamma S^{2}(e_{z})
\end{array}\right]
\end{equation}
where $I_{3\times3}$ is the $3\times3$ identity matrix. The characteristic
polynomial of the matrix $A_{2}$ is given by 
\begin{equation}
P(\lambda)=\left(\lambda^{2}+\alpha_{1}\lambda+\alpha_{2}\right)^{3}\lambda\left(\lambda-\gamma\right)^{2}.
\end{equation}
This polynomial has the following double real positive root 
\begin{equation}
\lambda=\gamma>0,
\end{equation}
It can be verified that the linearized system at $(0,0,2e_{z})$ admits
positive real eigenvalues. Hence the origin is almost globally asymptotically
stable.

This completes the proof of the theorem. 
\end{proof}

\subsection{Proof of Theorem~\ref{thm:RxRxS_exponential}}
\begin{proof}
The estimator has a unique stable equilibrium point defined by $\xi=\left(0,0,0\right)^{T}$.
The linearization of the error dynamics of (\ref{eq:RxRxS_error_dynamics_2})
around this point is defined in the tangent space to the state space
$\varUpsilon$, which happens to be $T\left(\varUpsilon\right)=\mathbb{R}^{3}\times\mathbb{R}^{3}\times\mathbb{R}^{2}$.
We define the state $\xi_{p}\overset{\Delta}{=}\left(z_{1},z'_{2},z_{2p}\right)$
where $z_{2p}\in\mathbb{R}^{2}$ is the orthogonal projection of $z_{2}$
on the horizontal plane. The linearized dynamics is given by $\dot{\xi}_{p}=A_{3}\xi_{p}$
with $A_{3}$ being a constant matrix having the form

\begin{equation}
A_{3}=\left[\begin{array}{ccccc}
-\alpha_{1}I_{3\times3} &  & -g_{0}I_{3\times3} &  & 0\\
\frac{\alpha_{2}}{g_{0}}I_{3\times3} &  & 0 &  & 0\\
0 &  & I_{2\times3} &  & -\gamma I_{2\times2}
\end{array}\right]
\end{equation}
where $I_{2\times3}=\left[\begin{array}{ccc}
1 & 0 & 0\\
0 & 1 & 0
\end{array}\right]$. The matrix $A_{3}$ is Hurwitz giving the stability of the linearized
dynamics. QED. 
\end{proof}
\bibliographystyle{unsrt}
\bibliography{biblio,biblio-local}

\end{document}